**Detecting Redundant Health Survey Questions Using Language-agnostic BERT Sentence Embedding (LaBSE)**

Sunghoon Kang, MS[a], Hyeoneui Kim, RN, MPH, PhD[b,c*], Hyewon Park, RN, BSN[b,d], Ricky Taira, PhD[e]

[a] College of Natural Sciences, Seoul National University, Seoul, Republic of Korea

[b] College of Nursing, Seoul National University, Seoul, Republic of Korea

[c] The Research Institute of Nursing Science, Seoul National University, Seoul, Republic of Korea

[d] Samsung Medical Center, Seoul, Republic of Korea

[e] The Department of Radiological Sciences, David Geffen School of Medicine, University of California Los Angeles, Los Angeles, CA, USA

* Corresponding Author

Hyeoneui Kim, RN, MPH, PhD

103 Daehak-ro, Jongno-gu, Seoul, 03080, Republic of Korea

Email: ifilgood@snu.ac.kr

Phone: +82) 02-740-8487

**Abstract**

**Objective**

The goal of this work was to compute the semantic similarity among publicly available health survey questions in order to facilitate the standardization of survey-based Person-Generated Health Data (PGHD).

**Methods**

We compiled various health survey questions authored in both English and Korean from the NIH CDE Repository, PROMIS, Korean public health agencies, and academic publications. Questions were drawn from various health lifelog domains. A randomized question pairing scheme was used to generate a Semantic Text Similarity (STS) dataset consisting of 1758 question pairs. Similarity scores between each question pair were assigned by two human experts. The tagged dataset was then used to build three classifiers featuring: Bag-of-Words, SBERT with BERT-based embeddings, and SBRET with LaBSE embeddings. The algorithms were evaluated using traditional contingency statistics.

**Results**

Among the three algorithms, SBERT-LaBSE demonstrated the highest performance in assessing question similarity across both languages, achieving an Area Under the Receiver Operating Characteristic (ROC) and Precision-Recall Curves of over 0.99. Additionally, it proved effective in identifying cross-lingual semantic similarities.

**Conclusion**

The SBERT-LaBSE algorithm excelled at aligning semantically equivalent sentences across both languages but encountered challenges in capturing subtle nuances and maintaining computational efficiency. Future research should focus on testing with larger multilingual datasets and on calibrating and normalizing scores across the health lifelog domains to improve consistency. This study introduces the SBERT-LaBSE algorithm for calculating semantic similarity across two languages, showing it outperforms BERT-based models and the Bag of Words approach, highlighting its potential to improve semantic interoperability of survey-based PGHD across language barriers.

# Introduction

Person-Generated Health Data (PGHD) is becoming increasingly important in managing individual health. PGHD encompasses health-related information that individuals create and collect outside traditional clinical environments, helping them monitor and manage their well-being [1,2]. Examples of PGHD include biometric data from wearable devices and self-reported information such as patient-reported outcomes (PROs). Given its potential for continuously capturing health insights beyond healthcare settings, there is growing interest in leveraging PGHD for both clinical care and health research [3-7]. However, the effective use of PGHD faces several challenges, including developing robust data management systems, ensuring data security, deploying it seamlessly into clinical workflows, and maintaining high data quality [6-8].

Standardizing survey-based Person-Generated Health Data (PGHD) is a critical step in enabling its broader use [9]. An important aspect of standardization is to identify redundancies in the form of semantic equivalencies. These may arise, since dependent upon the author, the clarity, tone, tense, directness, and formality of the language can be phrased differently for the same purposeful inquiry. For example, emotional symptoms might be captured by questions like, "Do you feel like withdrawing from family or friends?" or "I don't really want to talk to people around me." This variation makes identifying semantically equivalent questions—and thus standardizing survey-based PGHD—a complex task.

# Related Work

Efforts such as the Patient-Reported Outcomes Measurement Information System (PROMIS) and the National Institutes of Health (NIH) Common Data Elements (CDE) repository aim to provide standardized health survey questions. PROMIS, a consensus-based item bank designed for managing patient-reported outcomes (PROs), offers standardized measures that are applicable across various diseases and clinical settings [10-12]. These measures have helped healthcare providers manage patient symptoms, tailor treatments, and improve communication between patients and clinicians [13-16]. The NIH CDE repository, through metadata tagging, also plays a key role in standardizing data elements, including health surveys [17,18]. Both PROMIS and the CDE repository are essential for enhancing the interoperability of health data.

In practice, the deployment of PGHD acquisition applications requires survey questions be drawn from these established standardized resources. Data collected using questions outside of these resources still requires additional efforts to achieve standardization. While previous studies have explored ontology-mediated methods to identify semantically equivalent health questions [10,11], annotating each question with ontology concepts is labor-intensive and lacks scalability as such knowledge sources expand.

To address these challenges, we developed SPURT (Standardized PGHD Utilization Resources and Tools). SPURT supports the standardization and reuse of survey-based PGHD by identifying semantically equivalent questions, and facilitating the storage, retrieval, and sharing of this data. Unlike PROMIS and the NIH CDE Repository, SPURT annotates and stores health survey questions in both English and Korean, while also detecting semantically redundant questions. This ensures the use of consistent question formats whenever possible.

Technically, assessing semantic similarity between texts is a well-established yet widely applied method for managing text resources [19]. However, SPURT faces two unique challenges in its assigned task. First, it must effectively assess semantic similarities within or between two different languages, English and Korean. While multilingual embeddings can be used to address this challenge [20,21], they often perform less effectively for low-resource languages like Korean compared to high-resource languages such as English [22]. One common solution is to translate low-resource languages into high-resource ones before embedding, but this approach risks losing or distorting the original meaning [23,24]. The second challenge is ensuring computational efficiency for real-time semantic comparisons between questions. Calculating semantic similarity using large language models (LLMs) like BERT (Bidirectional Encoder Representations from Transformers) is computationally expensive, with a time complexity of $O(N!)$. For example, computing the similarity of approximately 10,000 sentence pairs can take around 65 hours using a V100 GPU [25]. Given that SPURT is designed to be a real-time, reactive data processing tool, achieving reasonable response times is crucial for its functionality.

This study presents the development of a novel algorithm for detecting redundant questions, addressing the challenges outlined above. The algorithm utilizes Sentence-BERT (SBERT), a variant of BERT designed for efficient sentence-level semantic similarity calculations [25], along with Language-agnostic BERT Sentence Embedding (LaBSE) [26] to enhance multilingual capability.

# Methods

## *Corpus Description: The Semantic Textual Similarity (STS) dataset*

A Semantic Textual Similarity (STS) dataset contains text pairs along with predefined similarity scores that quantify their semantic closeness [27-31]. This study developed an STS dataset to fine-tune pre-trained language models and evaluate our algorithms' performance in determining the semantic similarity between health-related questions.

We collected English and Korean questions from self-reported questionnaires covering the five health lifelog domains including diet, physical activity, living environment, stress management, and sleep. English questions (N=1,222) were sourced from the NIH CDE Repository, PROMIS, and academic publications, while Korean questions (N=963) were gathered from online resources provided by public health agencies and hospitals in Korea [17,32-35].

To build the STS dataset, we began by randomly selecting five seed questions from each of the five health lifelog domains in Korean, resulting in 25 seed questions. For each question, correspondingly similar questions for Korean were identified, resulting in 25 similar seed questions for each language. This correspondence of seed questions was performed in order to minimize the effects of semantic complexity on algorithm performance. We then randomly selected 30 comparison questions for each seed question, which yielded a total of 1500 question pairs (750 in each language).

The gold standard for semantic similarity between the question pairs was determined by two researchers with nursing backgrounds who independently scored the similarity of each question pair, following a standardized scoring protocol (Table 1). The agreement between the researchers, as measured by Cohen's kappa, varied by the health lifelog domains: 0.91 for diet, 0.72 for living environment, 0.83 for physical activity, 0.86 for sleep, and 1.0 for stress management, with an average Cohen's kappa of 0.86 across all the health lifelog domains.

Upon completion of this annotation process, we observed that the initial distribution of similarity scores was imbalanced, skewed heavily toward lower similarity scores. Only 2.3% of pairs received a score of 4, and 4.9% received a score of 3. To address this imbalance, we supplemented the dataset with an additional 117 English and 142 Korean question pairs from other sources, chosen to increase the frequency of semantically similar (i.e., higher scores) samples in the evaluation STS dataset. These additions brought the final evaluation set to include 820 question pairs (410 in each language), with the following distribution: 12.2%, 30.5%, 26.8%, and 30.5% for scores 4, 3, 2, and 1 respectively.

**Table 1.** Scoring protocol for semantic similarity.

| Score | Scoring Protocol | Examples |
|---|---|---|
| | | Seed question: *In the past month, have you ever had chest pain when you were not performing any physical activity?* |
| 4 | Minor differences in word choice from the seed question, but takes the same form of response | In the past month, have you had chest pain when you were not doing physical activity? |
| 3 | Share the same key topics, though some details may be added, altered, or omitted compared to the seed question. | Do you feel pain in your chest when you do physical activity? |
| 2 | The key topics are similar but more specific or general than that of the seed question. | Has your doctor ever said that you have a heart condition and that you should only perform physical activity recommended by a doctor? |
| 1 | Does not share the core topic from the seed question or belongs to a completely different health lifelog domain. | Have you done general conditioning exercises in the past 4 weeks? |

Using a similar procedure as described above, we compiled a second English STS dataset for the purpose of fine-tuning our pre-trained language models (see below). This fine-tuning dataset included 938 annotated English question pairs. The fine-tuning set had a distribution of 6.2% scoring 4, 14.0% scoring 3, 23.5% scoring 2, and 56.4% scoring 1.

In total, the STS dataset consisted of 1,758 question pairs, broken down into 820 for evaluation testing (410 English and 410 Korean) and 938 in English for classifier model refinement (see Supplement A). The process of constructing the STS dataset is illustrated in Fig 1.

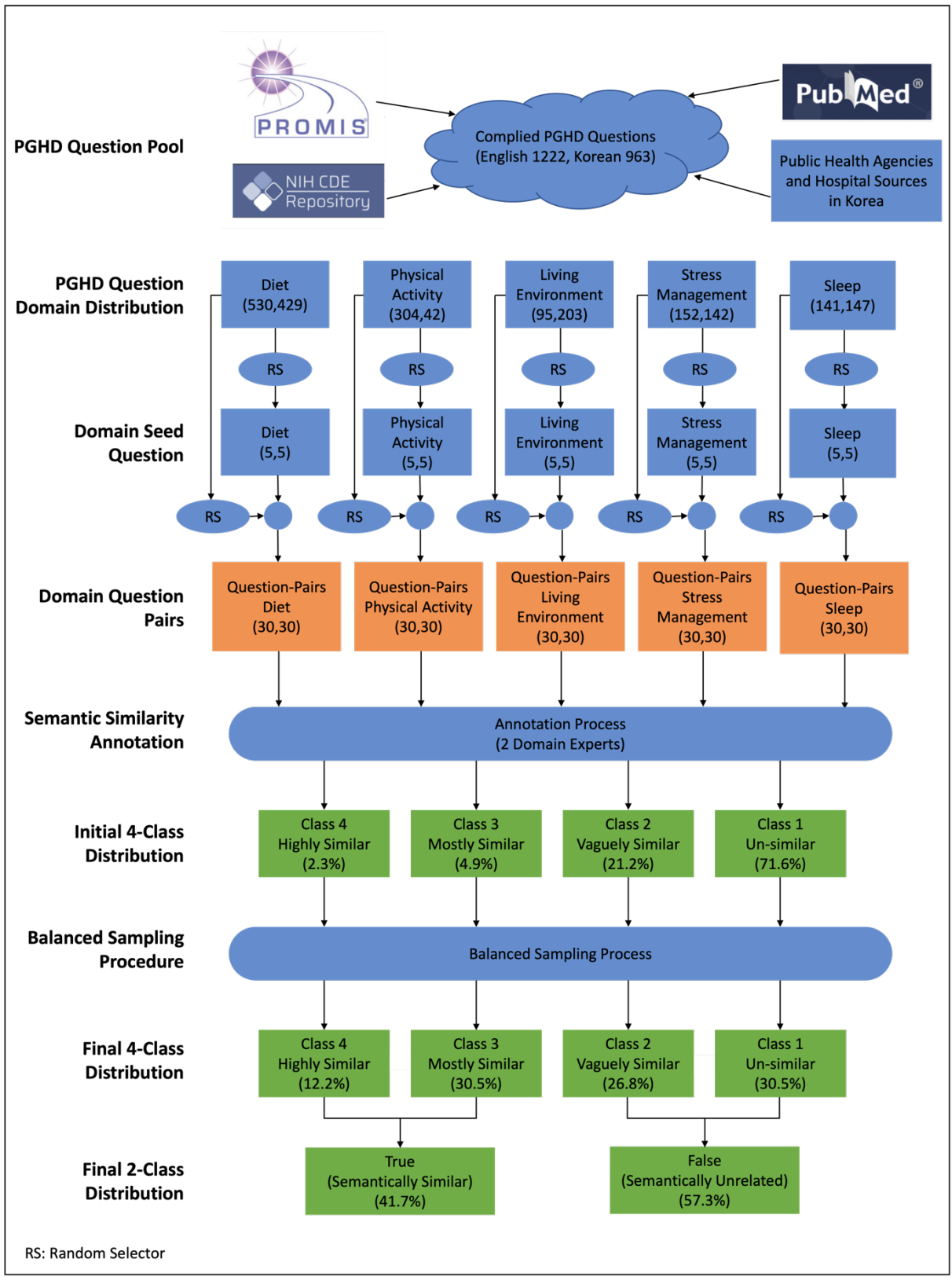

**Fig 1.** The process of preparing the STS dataset for fine-tuning and evaluation.

*Semantic similarity calculation algorithms*

We developed 3 classifiers to compare their performance capability for distinguishing the binary task of semantic similarity between STS question pairs. These included: i) a bag-of-words (BoW) model; ii) a Sentence-BERT with BERT-based embeddings (SBERT-BERT); and iii) a language-agnostic BERT Sentence Embedding model (LaBSE). Model fine-tuning and algorithm development were carried out using Python (version 3.11).

    a. Bag of Words (BoW) classifier

The BoW algorithm, a traditional language model that represents sentences by word frequency, serves as the baseline [36]. The BoW model's vocabulary was derived from the STS dataset, comprising 1,349 unique word forms after stop-word removal and lemmatization. Each sentence was represented as a 1,349-dimensional vector based on the vocabulary. Cosine similarity was used to calculate the semantic distance of question pairs. For Korean questions, translation into English was performed using the Google Translator API prior to similarity calculation [37].

    b. The SBERT-BERT algorithm

The SBERT-BERT large language model was derived from the pre-trained BERT-base model, which has 12 layers, a 768-dimensional hidden layer, and 12 attention heads [25]. SBERT-BERT supports only English. We fine-tuned the pre-trained SBERT-BERT model to optimize its performance for identifying semantic equivalency among health questions using the 938 English question pairs described above. The fine-tuning was performed with a batch size of 32, 8 epochs, and a learning rate of 2e-5, which were deemed optimal after testing various configurations. The AdamW optimizer was used for model optimization [38]. The fine-tuned SBERT-BERT algorithm was then evaluated using the test STS dataset of 410 English question pairs and 410 Korean question pairs. As previously stated, the Korean questions were translated into English using the Google Translator API, in order to execute the evaluation.

    c. The SBERT-LaBSE algorithm

The SBERT-LaBSE algorithm differs from SBERT-BERT in that it supports multiple languages within a single embedding space [26]. The pre-trained SBERT-LaBSE model was derived from the LaBSE model, which also consists of 12 layers, a 768-dimensional hidden layer, and 12 attention heads [26]. Fine-tuning was performed in the same manner as for SBERT-BERT. Unlike the other models, SBERT-LaBSE can assess the semantic similarity of English and Korean questions without requiring translation.

*Performance evaluation*

The performance of the similarity calculation algorithms was evaluated as a binary classification problem to simplify interpretation. The 4-point ordinal similarity scores from the STS dataset were converted into binary labels, where scores of 3 and 4 were categorized as 'similar' and scores of 1 and 2 as 'dissimilar.'

Optimal thresholds for predicting similarity were determined for the continuous similarity scores, which ranged from -1 to 1. Precision, recall, and F1-scores were calculated to assess algorithm performance, and the Area Under the Curve (AUC) for both the Receiver Operating Characteristics (ROC) and Precision-Recall (PR) curves were examined. The processes used by the three algorithms to calculate similarity are illustrated in Fig 2.

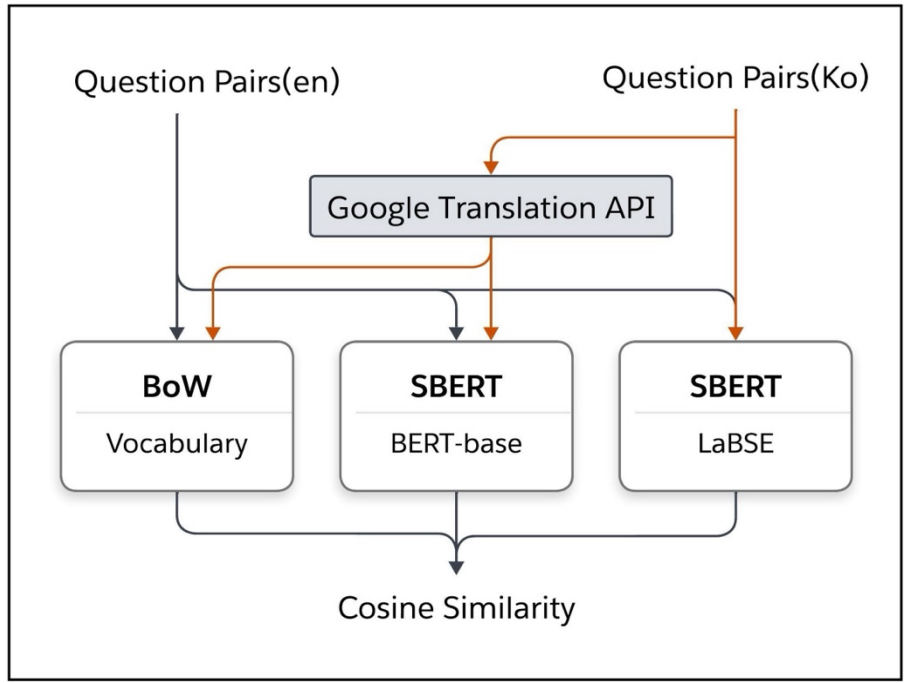

**Fig 2**. Similarity calculation with the three algorithms.

## Results

The performance of the three models for classifying similar vs. unsimilar question pairs when aggregating the five health lifelog domains is summarized in Table 2 and Fig 3. In the zero-shot trials (i.e., without the model refining stage), there were minimal differences in performance among the three algorithms for both English and Korean questions. All algorithms exhibited higher recall than precision in both languages. After fine-tuning, the SBERT-BERT algorithm showed a substantial improvement, particularly for English questions, in which the F1 score was raised from 0.65 to 0.96. For Korean questions, the improvement was moderate, with F1 score progressing from 0.68 to 0.73. In contrast, SBERT-LaBSE demonstrated significant improvements for both languages post fine-tuning. For English, F1 scores were raised from 0.66 to 0.98 while for Korean, F1 scores increased from 0.68 to 0.98. Fine tuning for both the SBERT-BERT and the SBERT-LaBSE models resulted in noticeable balanced performance between recall and precision.

Table 2. Performance Metrics for the three algorithms, combining the health lifelog domains.

| Language | Performance Metrics | Bag of Words | SBERT with pre-trained | | SBERT with fine-tuned | |
|---|---|---|---|---|---|---|
| | | | BERT-base | LaBSE | BERT-base | LaBSE |
| English question pairs (N=410) | Accuracy | 0.6112 | 0.6308 | 0.5917 | 0.9702 | 0.9853 |
| | Precision | 0.5279 | 0.5451 | 0.5111 | 0.9668 | 0.9818 |
| | Recall | 0.8161 | 0.7989 | 0.9253 | 0.9632 | 0.9839 |
| | F1 | 0.6411 | 0.6480 | 0.6585 | 0.9649 | 0.9828 |
| Korean question pairs (N=410) | Accuracy | 0.6610 | 0.6659 | 0.6878 | 0.7576 | 0.9839 |
| | Precision | 0.5732 | 0.5760 | 0.6054 | 0.6929 | 0.9818 |
| | Recall | 0.8057 | 0.8229 | 0.7714 | 0.7817 | 0.9806 |
| | F1 | 0.6698 | 0.6776 | 0.6784 | 0.7332 | 0.9812 |

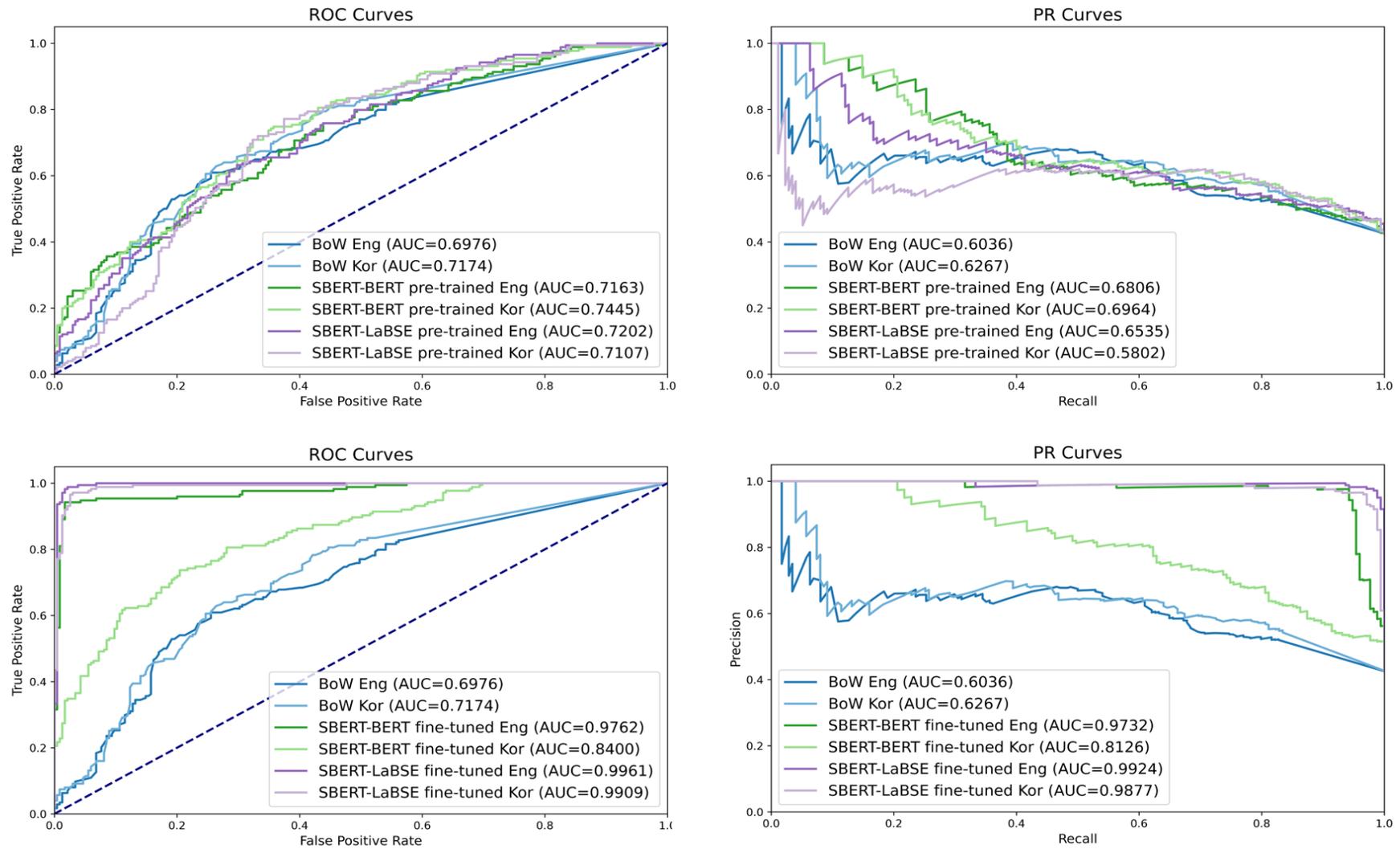

Fig 3. ROC and PR curves for pre-trained and fine-tuned embeddings on English and Korean questions, combining the health lifelog domains

**Table 3**. Performance Metrics of the SBERT based algorithms with the fine-tuned BERT and LaBSE models by the health lifelog domains.

| | Performance Metrics | English question pairs (N=410) | | | | | | Korean question pairs (N=410) | | | | | |
|---|---|---|---|---|---|---|---|---|---|---|---|---|---|
| | | DL (N=80) | HLE (N=80) | PA (N=80) | Sleep (N=85) | Stress (N=85) | All | DL (N=80) | HLE (N=80) | PA (N=80) | Sleep (N=85) | Stress (N=85) | All |
| BoW | Accuracy | 0.7215 | 0.7250 | 0.6625 | 0.4118 | 0.7176 | 0.6112 | 0.7250 | 0.8125 | 0.7250 | 0.5765 | 0.5882 | 0.6610 |
| | Precision | 0.7000 | 0.6383 | 0.5952 | 0.4118 | 0.6279 | 0.5279 | 0.6585 | 0.7941 | 0.6275 | 0.0000 | 0.5000 | 0.5732 |
| | Recall | 0.6176 | 0.8571 | 0.7143 | 1.0000 | 0.7714 | 0.8161 | 0.7714 | 0.7714 | 0.9143 | 0.0000 | 0.8286 | 0.8057 |
| | F1 | 0.6563 | 0.7317 | 0.6494 | 0.5833 | 0.6923 | 0.6411 | 0.7105 | 0.7826 | 0.7442 | 0.0000 | 0.6237 | 0.6698 |
| | ROC AUC | 0.7297 | 0.7457 | 0.6810 | 0.5820 | 0.7611 | 0.6976 | 0.7667 | 0.7937 | 0.7933 | 0.5937 | 0.6609 | 0.7174 |
| | PR AUC | 0.7250 | 0.6718 | 0.6301 | 0.5025 | 0.6834 | 0.6036 | 0.7394 | 0.7373 | 0.7519 | 0.4498 | 0.5985 | 0.6267 |
| SBERT with Fine-tuned BERT-base | Accuracy | 0.9646 | 0.9625 | 0.9800 | 0.9906 | 0.9835 | 0.9702 | 0.8525 | 0.8125 | 0.7025 | 0.7200 | 0.7882 | 0.7576 |
| | Precision | 0.9650 | 0.9502 | 0.9784 | 0.9836 | 0.9830 | 0.9668 | 0.8037 | 0.7391 | 0.6175 | 0.6036 | 0.7108 | 0.6929 |
| | Recall | 0.9529 | 0.9657 | 0.9771 | 0.9943 | 0.9771 | 0.9632 | 0.8800 | 0.8914 | 0.8629 | 0.9543 | 0.8229 | 0.7817 |
| | F1 | 0.9585 | 0.9571 | 0.9770 | 0.9887 | 0.9799 | 0.9649 | 0.8384 | 0.8062 | 0.7176 | 0.7376 | 0.7622 | 0.7332 |
| | ROC AUC | 0.9859 | 0.9698 | 0.9923 | 0.9929 | 0.9936 | 0.9867 | 0.9125 | 0.8563 | 0.7901 | 0.8411 | 0.8462 | 0.8412 |
| | PR AUC | 0.9858 | 0.9480 | 0.9925 | 0.9870 | 0.9918 | 0.9800 | 0.9008 | 0.7969 | 0.7640 | 0.8109 | 0.8244 | 0.8134 |
| SBERT with Fine-tuned LaBSE | Accuracy | 0.9848 | 0.9900 | 0.9875 | 0.9906 | 0.9906 | 0.9853 | 0.9775 | 0.9975 | 0.9850 | 0.9859 | 0.9835 | 0.9839 |
| | Precision | 0.9716 | 0.9889 | 0.9728 | 0.9944 | 0.9889 | 0.9818 | 0.9719 | 0.9944 | 0.9836 | 0.9775 | 0.9886 | 0.9818 |
| | Recall | 0.9941 | 0.9886 | 1.0000 | 0.9829 | 0.9886 | 0.9839 | 0.9771 | 1.0000 | 0.9829 | 0.9886 | 0.9714 | 0.9806 |
| | F1 | 0.9826 | 0.9885 | 0.9861 | 0.9884 | 0.9887 | 0.9828 | 0.9743 | 0.9972 | 0.9828 | 0.9829 | 0.9797 | 0.9812 |
| | ROC AUC | 0.9965 | 0.9929 | 0.9987 | 0.9989 | 0.9979 | 0.9968 | 0.9893 | 0.9976 | 0.9962 | 0.9971 | 0.9930 | 0.9951 |
| | PR AUC | 0.9964 | 0.9901 | 0.9984 | 0.9985 | 0.9975 | 0.9960 | 0.9872 | 0.9958 | 0.9947 | 0.9957 | 0.9927 | 0.9934 |

DL: Dietary Lifestyle, HLE: Human Living Environment, PA: Physical Activity

Table 3 presents the performance of the two SBERT algorithms across the five health lifelog domains. For all the health lifelog domains, the fine-tuned SBERT-BERT and SBERT-LaBSE algorithms demonstrated high performance on English questions, with ROC AUC and PR AUC values exceeding 0.95 and approaching 0.99. However, the SBERT-BERT algorithm struggled with the English-translated Korean questions, particularly in the Physical Activity domain. In contrast, the SBERT-LaBSE algorithm consistently delivered strong performance across all the health lifelog domains even for Korean questions.

Table 4 presents the optimal cut-off values for the three algorithms. The pre-trained SBERT-BERT and SBERT-LaBSE models showed considerable variation in cut-off values across the five health lifelog domains. However, after fine-tuning, these variations decreased, indicating that fine-tuning helped stabilize the algorithms. Despite this improvement, the SBERT-LaBSE algorithm still exhibited more variability in cut-off values across the health lifelog domains compared to SBERT-BERT, suggesting that further calibration may be required for SBERT-LaBSE. Supplement B provides example question pairs from each health lifelog domain, along with the similarity scores assigned by human reviewers and predicted by the three algorithms.

**Table 4**. Optimal cut-off for algorithms on BoW and pre-trained and fine-tuned SBERT-BERT and SBERT-LaBSE in each health lifelog domain.

| Language | Health lifelog Domain | Bag of Words | SBERT with pre-trained | | SBERT with fine-tuned | |
|---|---|---|---|---|---|---|
| | | | BERT-base | LaBSE | BERT-base | LaBSE |
| English question pairs (N=410) | DL | 0.2887 | 0.6274 | 0.5359 | 0.6349 | 0.6262 |
| | HLE | 0.1291 | 0.5369 | 0.3965 | 0.6151 | 0.6425 |
| | PA | 0.3162 | 0.3667 | 0.4822 | 0.6304 | 0.6202 |
| | Sleep | 0.0000 | 0.6790 | 0.2456 | 0.6617 | 0.6574 |
| | Stress | 0.1054 | 0.5817 | 0.3807 | 0.6359 | 0.5958 |
| | All | 0.1291 | 0.5816 | 0.3796 | 0.6278 | 0.6091 |
| Korean question pairs (N=410) | DL | 0.2887 | 0.5990 | 0.3103 | 0.5639 | 0.6568 |
| | HLE | 0.2582 | 0.5475 | 0.5603 | 0.5639 | 0.7138 |
| | PA | 0.1491 | 0.4778 | 0.6004 | 0.5639 | 0.6741 |
| | Sleep | 0.9354 | 0.4837 | 0.9215 | 0.5639 | 0.6849 |
| | Stress | 0.1091 | 0.6647 | 0.4481 | 0.5639 | 0.6586 |
| | All | 0.1336 | 0.5320 | 0.5753 | 0.5639 | 0.6531 |

Fig 4 illustrates that SBERT-LaBSE effectively determined semantic similarities between the two languages, with slightly better performance in identifying the semantic similarities of English questions relative to the Korean seed questions. The full results of the cross-language semantic similarity analysis are provided in Supplement C.

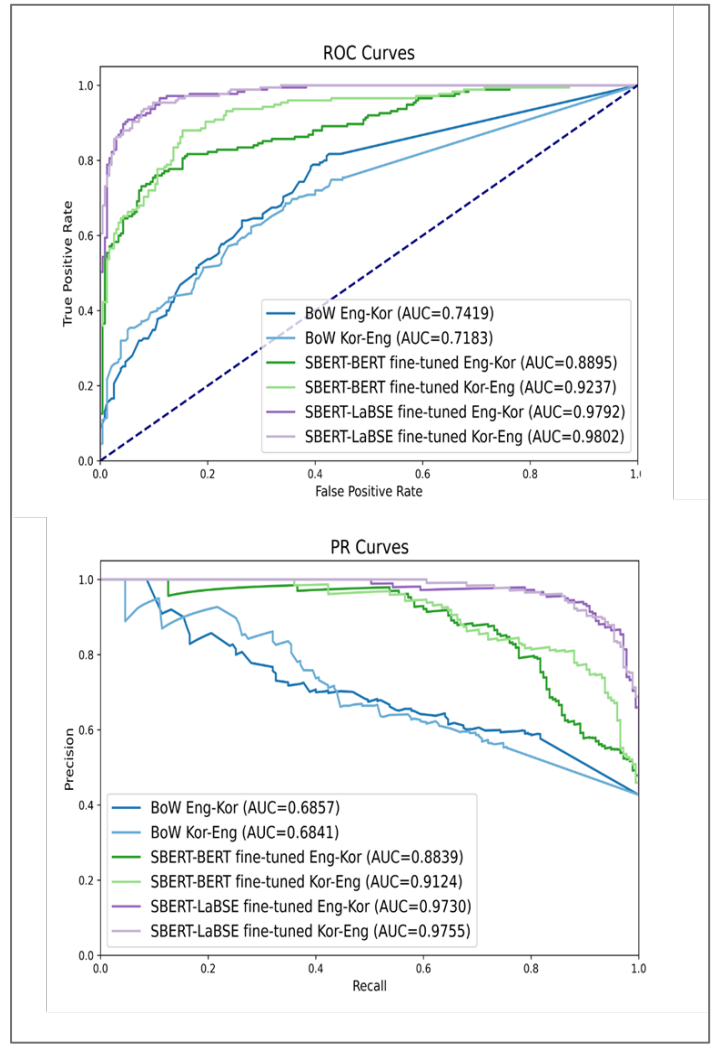

Fig 4. Performance of the cross-language semantic similarity determination.

## Discussion

This study demonstrates the utility of large language models for the purpose of determining semantic similarities among health questions in order to facilitate the standardization of survey-based health data. The results indicate that the fine-tuned SBERT algorithms were significantly more effective than the traditional BoW approach in identifying semantic similarities for both English and Korean questions. Furthermore, the SBERT-LaBSE algorithm demonstrated superior performance, particularly for Korean questions, suggesting that it is a more effective method than the SBERT-BERT algorithm, which relies on English translation, for assessing semantic similarity in non-English texts.

The SBERT-LaBSE algorithm's success with Korean questions can be attributed to its structural design and the limitations of language translation. Structurally, LaBSE aligns semantically equivalent words or sentences from different languages into a unified embedding space, preserving semantic consistency across languages. This allows for more accurate semantic similarity assessments. In contrast, the SBERT-BERT algorithm's lower performance with Korean questions may be due to meaning loss or distortion during translation, which disrupts semantic comparisons between languages [23,24]. While previous studies have noted that LaBSE may struggle with subtle, sentence-level nuances, limiting its performance in fine-grained similarity tasks [39], our study found that the SBERT-LaBSE algorithm effectively captured the meanings in both English and Korean sentences, outperforming the SBERT-BERT model. However, this finding should be validated with a larger, more diverse dataset that includes a broader range of syntactic features.

When implemented in the SPURT system with 1,835 questions in the comparison space, the SBERT-LaBSE algorithm evaluated the similarity of a new question in just 0.03 seconds. This was achieved on a Naver Cloud Platform server with 8GB RAM and no GPU [40]. Despite its impressive performance, LaBSE's 440 million parameters—four times that of BERT-base—make it a resource-intensive option, potentially increasing costs for complex tasks. This resource demand may limit its applicability on resource-constrained devices such as mobile platforms [41].

This study has some limitations. First, the cut-off values for the similarity scores were not uniformly calibrated across the five health lifelog domains, leading to inconsistencies in how similarity scores were interpreted. For example, the SBERT-LaBSE algorithm assigned a similarity score of 0.7 to the dietary question pair "I've binge eaten" and "Do you ever overeat?" and identified them as similar. On the other hand, the algorithm correctly identified the human living environment questions "Have you moved in the past 5 years?" and "In the last 5 years, the number of people in this community has?" dissimilar while assigning the same similarity score of 0.7 to the pair. These inconsistencies may impact the accurate interpretation of similarity scores, highlighting the need for future work to focus on calibrating and normalizing scores across the health lifelog domains to ensure greater consistency.

Second, the current evaluation was conducted on a relatively small set of English and Korean question pairs. Future studies should explore the feasibility of applying the SBERT-LaBSE algorithm to a broader range of sentence types from diverse domains. Additionally, by incorporating texts from more diverse languages, future research can investigate the algorithm's potential to overcome language barriers and facilitate semantic interoperability. As one potential area for future research, a comparative analysis of the SBERT-LaBSE algorithm with generative large language models could be explored. Exploring their performance in comparison to SBERT-LaBSE in multilingual contexts may provide additional insights for standardizing survey-based health data.

## Conclusion

This study presents evidence for the potential for utilizing large language models for identifying semantic redundancy within survey-based Person-Generated Health Data (PGHD) collections. We demonstrated that the SBERT-LaBSE algorithm, in particular, provided excellent classification for calculating semantic similarity across diverse question formats in two languages. The findings demonstrate that SBERT-LaBSE outperforms both BERT-based algorithms and the traditional bag-of-words (BoW) approach in both languages underscoring its potential to enhance the semantic interoperability of PGHD across language barriers.

**Acknowledgments**

We are thankful to our colleagues, Eunyeong Lim and Jeongha Kim, for their help with the STS dataset preparation.

39. Wang W, Chen G, Wang H, Han Y, Chen Y. Multilingual sentence transformer as a multilingual word aligner. *Findings of the Association for Computational Linguistics: EMNLP*; 2022.

40. Naver Cloud Platform. Accessed October 14, 2024. Available from: https://www.ncloud.com/product/compute/server#detail

41. Mao Z, Nakagawa T. LEALLA: Learning lightweight language-agnostic sentence embeddings with knowledge distillation. *Proceedings of the Conference of the European Chapter of the Association for Computational Linguistics*; 2023.

**Supplement A.** Distribution of Scores for Question Pairs in each language

**Table A-1.** Score Distribution of final STS dataset for evaluation

| score | English Question Pairs (N=410) | | | | | | Korean Question Pairs (N=410) | | | | | |
|---|---|---|---|---|---|---|---|---|---|---|---|---|
| | DL (N=80) | HLE (N=80) | PA (N=80) | Sleep (N=85) | Stress (N=85) | All | DL (N=80) | HLE (N=80) | PA (N=80) | Sleep (N=85) | Stress (N=85) | All |
| 4 | 10 (12.5%) | 10 (12.5%) | 10 (12.5%) | 10 (11.8%) | 10 (11.8%) | 50 (12.2%) | 10 (12.5%) | 10 (12.5%) | 10 (12.5%) | 10 (11.8%) | 10 (11.8%) | 50 (12.2%) |
| 3 | 25 (31.3%) | 25 (31.3%) | 25 (31.3%) | 25 (29.4%) | 25 (29.4%) | 125 (30.5%) | 25 (31.3%) | 25 (31.3%) | 25 (31.3%) | 25 (29.4%) | 25 (29.4%) | 125 (30.5%) |
| 2 | 20 (25.0%) | 20 (25.0%) | 20 (25.0%) | 25 (29.4%) | 25 (29.4%) | 110 (26.8%) | 20 (25.0%) | 20 (25.0%) | 20 (25.0%) | 25 (29.4%) | 25 (29.4%) | 110 (26.8%) |
| 1 | 25 (31.3%) | 25 (31.3%) | 25 (31.3%) | 25 (29.4%) | 25 (29.4%) | 125 (30.5%) | 25 (31.3%) | 25 (31.3%) | 25 (31.3%) | 25 (29.4%) | 25 (29.4%) | 125 (30.5%) |

**Table A-2.** Score Distribution of final STS dataset for fine-tuning

| score | English Question Pairs (N=938) | | | | | |
|---|---|---|---|---|---|---|
| | DL (N=182) | HLE (N=191) | PA (N=186) | Sleep (N=172) | Stress (N=207) | All |
| 4 | 11 (6.0%) | 10 (5.2%) | 12 (6.5%) | 15 (8.7%) | 10 (4.8%) | 58 (6.2%) |
| 3 | 26 (14.3%) | 28 (14.7%) | 34 (18.3%) | 15 (8.7%) | 28 (13.5%) | 131 (14.0%) |
| 2 | 68 (37.4%) | 32 (16.8%) | 58 (31.2%) | 35 (20.3%) | 27 (13.0%) | 220 (23.5%) |
| 1 | 77 (42.3%) | 121 (63.4%) | 82 (44.1%) | 107 (62.2%) | 142 (68.6%) | 529 (56.4%) |

**Table A-3.** Percent Score Distribution of final STS dataset

| score | English Question Pairs (N=1348) | | | | | | Korean Question Pairs (N=410) | | | | | |
|---|---|---|---|---|---|---|---|---|---|---|---|---|
| | DL (N=262) | HLE (N=271) | PA (N=266) | Sleep (N=257) | Stress (N=292) | All | DL (N=80) | HLE (N=80) | PA (N=80) | Sleep (N=85) | Stress (N=85) | All |
| 4 | 21 (8.0%) | 20 (7.4%) | 22 (8.3%) | 25 (9.7%) | 20 (6.9%) | 108 (8.0%) | 10 (12.5%) | 10 (12.5%) | 10 (12.5%) | 10 (11.8%) | 10 (11.8%) | 50 (12.2%) |
| 3 | 51 (19.5%) | 53 (19.6%) | 59 (22.2%) | 40 (15.6%) | 53 (18.1%) | 256 (19.0%) | 25 (31.3%) | 25 (31.3%) | 25 (31.3%) | 25 (29.4%) | 25 (29.4%) | 125 (30.5%) |
| 2 | 88 (33.6%) | 52 (19.2%) | 78 (29.3%) | 60 (23.4%) | 52 (17.8%) | 330 (24.5%) | 20 (25.0%) | 20 (25.0%) | 20 (25.0%) | 25 (29.4%) | 25 (29.4%) | 110 (26.8%) |
| 1 | 102 (38.9%) | 146 (53.9%) | 107 (40.2%) | 132 (51.4%) | 167 (57.2%) | 654 (48.5%) | 25 (31.3%) | 25 (31.3%) | 25 (31.3%) | 25 (29.4%) | 25 (29.4%) | 125 (30.5%) |

**Supplement B. Example question pairs with the scores from human review and predictions from the three algorithms.**

**Table B-1.** Question pairs between English and English in the Physical Activity domain.

S: Similar, D: Dissimilar

**Seed question:** In the past month, have you ever had chest pain when you were not performing any physical activity?

| Question | score | BoW | SBERT-BERT | | SBERT-LaBSE | |
| --- | --- | --- | --- | --- | --- | --- |
| | | | zero-shot | fine-tuned | zero-shot | fine-tuned |
| In the past month, have you had chest pain when you were not doing physical activity? | 4 | S | S | S | S | S |
| In the last month, have you experienced chest pain even when you weren't exercising? | 4 | S | S | S | S | S |
| Do you feel pain in your chest when you do physical activity? | 3 | S | D | S | S | S |
| Do you feel pain in your chest when you perform physical activity? | 3 | S | D | S | S | S |
| Have you ever felt pain in your chest when you exercise? | 3 | S | S | S | S | S |
| Do you have chest pain when you exercise? | 3 | S | D | S | S | S |
| Do you often have chest pain and suffer from it? | 3 | S | D | S | D | S |
| Has your doctor ever said that you have a heart condition and that you should only perform physical activity recommended by a doctor? | 2 | S | D | D | D | D |
| In the past month, how often did you have problems with participating in sports activity or exercise? | 2 | S | S | D | S | D |
| Were you sick last week, or did anything prevent you from doing your normal physical activities? | 2 | D | S | D | S | D |
| During the past 4 weeks, how much did physical health problems limit your usual physical activities (such as walking or climbing stairs)? | 2 | S | S | D | D | D |
| I'm in too much pain to exercise | 2 | D | D | D | D | D |
| Have you done general conditioning exercises in the past 4 weeks? | 1 | D | D | D | S | D |
| Have you done moderate to heavy strength training in the past 4 weeks? | 1 | D | D | D | S | D |

| Seed question: In the past month, have you ever had chest pain when you were not performing any physical activity? | | | | | | |
|---|---|---|---|---|---|---|
| Question | score | BoW | SBERT-BERT | | SBERT-LaBSE | |
| | | | zero-shot | fine-tuned | zero-shot | fine-tuned |
| Have you walked leisurely for exercise or pleasure in the past 4 weeks? | 1 | D | S | D | S | D |
| Have you done yoga or Tai-chi in the past 4 weeks? | 1 | D | D | D | D | D |
| I rarely or never do any physical activities. | 1 | S | D | D | D | D |

**Table B-2.** Question pairs between Korean and Korean in the Stress Management domain.

S: Similar, D: Dissimilar

| Seed question: 일상적인 것이 아닌 사건들(범죄, 자연재해, 우발사고, 이사 등)로 인한 압박감의 정도는? | | | | | | |
|---|---|---|---|---|---|---|
| What is the degree of pressure from cases that are not everyday events (crime, natural disasters, accident accidents, moving, etc.)? | | | | | | |

| Question | score | BoW | SBERT-BERT | | SBERT-LaBSE | |
|---|---|---|---|---|---|---|
| | | | zero-shot | fine-tuned | zero-shot | fine-tuned |
| 최근 한 달 동안 우발사고로 인해 받은 스트레스의 정도는 어느 정도였나요? (What was the degree of stress that was caused by contingent accidents in recent months?) | 4 | S | S | S | S | S |
| 최근 한 달 동안 범죄와 같은 일상적이지 않은 사건으로 인해 느낀 스트레스의 정도는 어느 정도였나요? (What was the degree of stress that you felt because of a routine incident like crime in recent months?) | 4 | D | S | S | S | S |
| 지난 한 달 동안 일상적이지 않은 사건들로 인해 스트레스를 자주 받았었나요? (Have you been stressed often for the last month?) | 3 | D | D | D | S | S |
| 지난 한 달 동안 예상치 못한 일로 인해 스트레스를 받은 적이 얼마나 자주 있었나요? (How often have you been stressed for unexpected things over the past month?) | 3 | D | D | S | S | S |
| 지난 한 달 동안 자신의 범죄나 사고로 인해 스트레스를 받은 적이 얼마나 자주 있었나요? (How often have you been stressed by your crimes or accidents over the past month?) | 3 | S | D | S | D | S |
| 지난 한 달 동안 일상적이지 않은 일로 인해 화가 난 적이 얼마나 자주 있었나요? (How often have you been angry for the last month?) | 3 | D | D | D | D | S |
| 지난 한 달 동안 갑작스러운 사건들로 인해 스트레스를 받은 적이 얼마나 자주 있었나요? (How often have you been stressed by sudden events over the past month?) | 3 | D | D | S | S | S |
| (자신 혹은 타인의) 병이나 상해에 의한 압박감의 정도는? (What is the degree of pressure caused by illness or injury?) | 2 | D | S | S | D | D |

**Seed question:** 일상적인 것이 아닌 사건들(범죄, 자연재해, 우발사고, 이사 등)로 인한 압박감의 정도는?

What is the degree of pressure from cases that are not everyday events (crime, natural disasters, accident accidents, moving, etc.)?

| Question | score | BoW | SBERT-BERT | | SBERT-LaBSE | |
|---|---|---|---|---|---|---|
| | | | zero-shot | fine-tuned | zero-shot | fine-tuned |
| 대인관계의 변화(사망, 출생, 결혼, 이혼 등)로 인한 압박감의 정도는? (What is the degree of pressure from the change of interpersonal relationships (death, birth, marriage, divorce, etc.)?) | 2 | D | S | S | S | D |
| (자신이 원했든지 아니든지 간에)일, 직업 및 학교와 관계된 압박감의 정도는? (What is the degree of pressure related to work, job and school?) | 2 | D | D | D | D | D |
| 지난 일주간 전반적으로 느끼는 압박감의 정도는? (What is the degree of pressure you feel overall in the past week?) | 2 | D | D | D | D | D |
| 일상생활 중에 일어나는 사소한 변화(동회, 우체국 등을 찾아가는 일 등) 때문에 생기는 압박감의 정도는(만약 일상생활중 변화가 없다면 그것(권태)때문에 생기는 압박감의 정도를 표기하십시요)? (The degree of pressure from the minor changes that occur in everyday life (such as a club, a post office, etc.) is the degree of pressure (if there is no change in everyday life, indicate the degree of pressure from it (boredom))?) | 2 | S | S | S | S | D |
| 내 업무는 창의력을 필요로 한다. (My work requires creativity.) | 1 | D | D | D | D | D |
| 내 업무를 수행하기 위해서는 높은 수준의 기술이나 지식이 필요하다. (In order to perform my work, high levels of technology or knowledge are required.) | 1 | D | S | D | D | D |
| 권위자(윗사람, 직장상관 등)와 맞서게 되었다. (He was confronted with the authority (superior, workplace, etc.).) | 1 | D | D | D | D | D |
| 지난 한 달간 어느 정도로 경험했는지 오래 생각하시지 말고 최대한 빨리 응답해주세요. 2. 다른 사람으로부터 비난받거나 심판받는다고 느껴진다. (Don't think about how long you have experienced in the past month, but as soon as possible.2. It feels like being criticized or judged by others.) | 1 | D | D | D | D | D |
| 제시간에 맞추기 위해 서둘러야 했다. (I had to hurry to match the time.) | 4 | S | S | S | S | S |

**Table B-3.** Question pairs between English and Korean in the Living Environment domain.

S: Similar, D: Dissimilar

| Seed question: Do you have any complaints about your housing and its environment from those mentioned below: noise from neighbors and from outside ||||||||
|---|---|---|---|---|---|---|
| Question | score | BoW | SBERT-BERT || SBERT-LaBSE ||
| | | | zero-shot | fine-tuned | zero-shot | fine-tuned |
| 귀 댁이 현재 살고 계신 주거환경에 어느 정도 만족하십니까? 가장 많이 이용하는 시설이나 장소를 기준으로 응답해주십시오. 11) 자동차 경적, 집주변의 소음 정도 (How satisfied you are in your living environment? Please respond based on the most common facilities or places. 11) Car horn, degree of noise around the house) | 4 | S | D | S | S | S |
| 귀 댁이 현재 거주하는 주택의상태에 대해 평가해주십시오. 시설이 없는 경우는 불량으로 표기해주십시오. 6-1) 주택 외부 소음(차량 경적, 공사장 소음 등) (Please evaluate the status of your home. If there is no facility, please mark it in bad. 6-1) External noise outside housing (vehicle horn, construction noise, etc.)) | 4 | S | S | S | S | S |
| 쾌적한 주거환경 조성을 위해 어떠한 부분이 개선이 필요하다고 생각하십니까? 2) 외부 소음 (Do you think that needs improvement to create a pleasant residential environment? 2) External noise) | 3 | S | S | S | D | S |
| 현재 거주하는 주택의 외부 소음 때문에 생활하기가 어렵다. (It is difficult to live because of the external noise of the current house.) | 3 | S | S | S | S | S |
| 현재 거주하는 주택은 외부 소음이 거의 없는 편이다. (Currently, houses living in the current houses are rarely external noise.) | 3 | S | D | D | D | S |
| 현재 거주하는 주택은 외부/내부 소음이 심한 편이다. (Currently, houses that live in the current and internal noise are severe.) | 3 | S | D | S | S | S |
| 쾌적한 주거환경 조성을 위해 외부 소음 관리가 추가로 더 필요하다. (External noise management is needed to create a pleasant residential environment.) | 3 | S | S | S | D | S |
| 현재 거주공간에 대한 질문입니다. 아래 항목에 대해 느끼시는 정도를 해당 번호에 표시하여 주십시오. 6) 주택 실내가 좁다/넓다 (This is a question about the current living space. Please display the degree of feeling about the items below. 6) The room is narrow/wide) | 2 | D | D | D | D | D |

| Seed question: Do you have any complaints about your housing and its environment from those mentioned below: noise from neighbors and from outside | | | | | | |
|---|---|---|---|---|---|---|
| Question | score | BoW | SBERT-BERT | | SBERT-LaBSE | |
| | | | zero-shot | fine-tuned | zero-shot | fine-tuned |
| 현재 거주공간에 대한 질문입니다. 아래 항목에 대해 느끼시는 정도를 해당 번호에 표시하여 주십시오. 3) 전반적인 환경이 어둡다/밝다 <br> (This is a question about the current living space. Please display the degree of feeling about the items below. 3) The overall environment is dark/bright) | 2 | S | S | D | S | D |
| 현재 거주공간에 대한 질문입니다. 아래 항목에 대해 느끼시는 정도를 해당 번호에 표시하여 주십시오. 9) 주택 내 여름철 덥다/시원하다 <br> (This is a question about the current living space. Please display the degree of feeling about the items below. 9) Hot summer in the house/cool) | 2 | D | D | D | S | D |
| 현재 거주공간에 대한 질문입니다. 아래 항목에 대해 느끼시는 정도를 해당 번호에 표시하여 주십시오. 8) 주택 내 겨울철 춥다/따뜻하다 <br> (This is a question about the current living space. Please display the degree of feeling about the items below. 8) It is cold in winter in the house/warm) | 2 | D | D | D | D | D |
| 귀 댁이 현재 살고 계신 주거환경에 어느 정도 만족하십니까? 가장 많이 이용하는 시설이나 장소를 기준으로 응답해주십시오. 10) 치안 및 범죄 등 방범 상태 <br> (How satisfied you are in your living environment? Please respond based on the most common facilities or places. 10) security and crime status such as security and crime) | 2 | D | D | D | D | D |
| 귀하는 향후 10 년 이후에도 현재 자치구에서 거주하고 싶으십니까? <br> (Do you want to live in autonomous districts after the next 10 years?) | 1 | D | D | D | D | D |
| 귀 댁의 거주 위치는 어디에 해당됩니까? (Where is the location of your home?) | 1 | D | D | D | D | D |
| 이전 거주 주택 유형이 어떻게 되십니까? (What is the type of housing in the previous residence?) | 1 | S | D | D | D | D |
| 귀 댁의 점유형태는 어디에 해당됩니까? (Where is the occupancy of your house?) | 1 | D | D | D | D | D |
| 귀하는 현재 자치구에서 사신 지 몇 년이 되셨습니까? 총 거주기간을 말씀해 주십시오. <br> (How many years have you been living in autonomous districts? Please tell us the total residence period.) | 1 | D | D | D | D | D |

**Table B-4.** Question pairs between Korean and English in the Diet domain.

<div align="right">S: Similar, D: Dissimilar</div>

| Seed question: 억제할 수 없이 폭식을 한 적이 있다. (I have been binge eating without suppressing) | | | | | | |
|---|---|---|---|---|---|---|
| Question | score | BoW | SBERT-BERT | | SBERT-LaBSE | |
| | | | zero-shot | fine-tuned | zero-shot | fine-tuned |
| Have you ever experienced a loss of control while eating an unusually large amount of food, feeling unable to stop or manage what you were eating? | 4 | S | S | S | D | S |
| Have you had instances where you ate significantly more food than usual and felt you couldn't control your eating habits? | 4 | D | D | S | D | S |
| How many TIMES per week on average over the past 3 MONTHS have you eaten an unusually large amount of food and experienced a loss of control? | 3 | D | D | D | D | S |
| How many DAYS per week on average over the past 6 MONTHS have you eaten an unusually large amount of food and experienced a loss of control? | 3 | D | D | D | D | S |
| Feel very upset about your uncontrollable overeating or resulting weight gain? | 3 | D | D | D | D | S |
| How often do you find yourself engaging in binge eating behaviors? | 3 | S | S | S | D | S |
| Do you feel a loss of control while binge eating? | 3 | S | D | D | D | S |
| Eat large amounts of food when you didn't feel physically hungry? | 2 | D | S | D | D | D |
| During the past 6 months have there been times when you felt you have eaten what other people would regard as an unusually large amount of food (e.g. a quart of ice cream) given the circumstances? | 2 | D | D | D | D | D |
| Eat until you felt uncomfortably full? | 2 | D | D | D | D | D |
| Eating disorder | 2 | S | D | D | D | D |
| How often do you find yourself eating excessively, regardless of hunger cues? | 2 | S | S | S | D | D |
| I fear I may start choking when I eat food. | 1 | D | D | D | D | D |
| Has food intake declined over the past 3 months due to loss of appetite, digestive problems, chewing or swallowing difficulties? | 1 | D | D | D | D | D |
| Overall, when you think about the foods you ate over the past 12 months, would you say your diet was high, medium, or low in fat? | 1 | D | D | D | D | D |
| I don't enjoy eating anymore. | 1 | S | D | D | D | D |

| Seed question: 억제할 수 없이 폭식을 한 적이 있다. (I have been binge eating without suppressing) | | | | | | |
|---|---|---|---|---|---|---|
| Question | score | BoW | SBERT-BERT | | SBERT-LaBSE | |
| | | | zero-shot | fine-tuned | zero-shot | fine-tuned |
| Hyperorality/food fads: Has s/he been drinking or eating excessively anything in sight, or developing food fads, a sweet tooth, eating bananas or cookies excessively, or even putting objects in his/her mouth, or has s/he always had a large appetite and the eating habits have not changed? Has s/he lost table manners? | 1 | S | S | D | D | D |

**Supplement C.** Performance metrics in the cross-language semantic similarity analysis

**Table C-1**. Performance metrics on question pairs with English and Korean seed questions.

| Performance Metrics | English-Korean question pairs (N=410) | | | Korean-English question pairs (N=410) | | |
|---|---|---|---|---|---|---|
| | Bag of Words | SBERT with fine-tuned | | Bag of Words | SBERT with fine-tuned | |
| | | BERT-base | LaBSE | | BERT-base | LaBSE |
| Accuracy | 0.6780 | 0.8429 | 0.9317 | 0.6463 | 0.8488 | 0.9268 |
| Precision | 0.5885 | 0.8358 | 0.9298 | 0.5647 | 0.7703 | 0.8962 |
| Recall | 0.8171 | 0.7886 | 0.9086 | 0.7486 | 0.9200 | 0.9371 |
| F1 | 0.6842 | 0.8110 | 0.9191 | 0.6437 | 0.8385 | 0.9162 |
| ROC AUC | 0.7419 | 0.8905 | 0.9792 | 0.7183 | 0.9318 | 0.9802 |
| PR AUC | 0.6857 | 0.8911 | 0.9730 | 0.6841 | 0.9144 | 0.9755 |

**Table C-2**. Performance Metrics of the SBERT based algorithms with the fine-tuned BERT and LaBSE models by the health lifelog domains

| | Performance Metrics | English-Korean question pairs (N=410) | | | | | | Korean-English question pairs (N=410) | | | | | |
|---|---|---|---|---|---|---|---|---|---|---|---|---|---|
| | | DL (N=80) | HLE (N=80) | PA (N=80) | Sleep (N=85) | Stress (N=85) | All | DL (N=80) | HLE (N=80) | PA (N=80) | Sleep (N=85) | Stress (N=85) | All |
| BoW | Accuracy | 0.6875 | 0.8375 | 0.6875 | 0.6118 | 0.7176 | 0.6780 | 0.6250 | 0.7250 | 0.7750 | 0.4118 | 0.7294 | 0.6463 |
| | Precision | 0.5962 | 0.7750 | 0.6190 | 0.5208 | 0.6341 | 0.5885 | 0.5490 | 0.6444 | 0.7931 | 0.4118 | 0.6667 | 0.5647 |
| | Recall | 0.8857 | 0.8857 | 0.7429 | 0.7143 | 0.7429 | 0.8171 | 0.8000 | 0.8286 | 0.6571 | 1.0000 | 0.6857 | 0.7486 |
| | F1 | 0.7126 | 0.8267 | 0.6753 | 0.6024 | 0.6842 | 0.6842 | 0.6512 | 0.7250 | 0.7188 | 0.5833 | 0.6761 | 0.6437 |
| | ROC AUC | 0.7394 | 0.8984 | 0.7048 | 0.6769 | 0.7237 | 0.7419 | 0.6613 | 0.7683 | 0.7683 | 0.6449 | 0.7263 | 0.7183 |
| | PR AUC | 0.6744 | 0.8912 | 0.6594 | 0.6503 | 0.6394 | 0.6857 | 0.6215 | 0.7281 | 0.7445 | 0.6697 | 0.6919 | 0.6841 |
| SBERT with Fine-tuned BERT-base | Accuracy | 0.8700 | 0.9025 | 0.7925 | 0.8141 | 0.9176 | 0.8429 | 0.8500 | 0.8875 | 0.8625 | 0.8706 | 0.9647 | 0.8488 |
| | Precision | 0.8155 | 0.8956 | 0.8151 | 0.8287 | 0.9525 | 0.8358 | 0.7949 | 0.9063 | 0.7857 | 0.8529 | 0.9706 | 0.7703 |
| | Recall | 0.9143 | 0.8800 | 0.6971 | 0.6914 | 0.8457 | 0.7886 | 0.8857 | 0.8286 | 0.9429 | 0.8286 | 0.9429 | 0.9200 |
| | F1 | 0.8608 | 0.8876 | 0.7453 | 0.7537 | 0.8947 | 0.8110 | 0.8378 | 0.8657 | 0.8571 | 0.8406 | 0.9565 | 0.8385 |
| | ROC AUC | 0.9236 | 0.9468 | 0.8319 | 0.8398 | 0.9322 | 0.8905 | 0.9067 | 0.9473 | 0.9295 | 0.9251 | 0.9931 | 0.9318 |
| | PR AUC | 0.9125 | 0.9411 | 0.8491 | 0.8352 | 0.9382 | 0.8911 | 0.8802 | 0.9256 | 0.9194 | 0.9218 | 0.9911 | 0.9144 |
| SBERT with Fine-tuned LaBSE | Accuracy | 0.9000 | 1.0000 | 0.9625 | 0.9294 | 0.9765 | 0.9317 | 0.9375 | 0.9875 | 0.9875 | 0.8824 | 0.9765 | 0.9268 |
| | Precision | 0.8857 | 1.0000 | 0.9211 | 0.8718 | 0.9714 | 0.9298 | 0.9412 | 0.9722 | 1.0000 | 0.8049 | 1.0000 | 0.8962 |
| | Recall | 0.8857 | 1.0000 | 1.0000 | 0.9714 | 0.9714 | 0.9086 | 0.9143 | 1.0000 | 0.9714 | 0.9429 | 0.9429 | 0.9371 |
| | F1 | 0.8857 | 1.0000 | 0.9589 | 0.9189 | 0.9714 | 0.9191 | 0.9275 | 0.9859 | 0.9855 | 0.8684 | 0.9706 | 0.9162 |
| | ROC AUC | 0.9486 | 1.0000 | 0.9924 | 0.9703 | 0.9806 | 0.9792 | 0.9778 | 0.9968 | 0.9975 | 0.9543 | 0.9954 | 0.9802 |
| | PR AUC | 0.9311 | 1.0000 | 0.9900 | 0.9554 | 0.9708 | 0.9730 | 0.9744 | 0.9958 | 0.9971 | 0.9387 | 0.9942 | 0.9755 |

DL: Dietary Lifestyle, HLE: Human Living Environment, PA: Physical Activity